\documentclass[conference]{IEEEtran}
\IEEEoverridecommandlockouts
\usepackage{cite}
\usepackage{amsmath,amssymb,amsfonts}
\usepackage{algorithmic}
\usepackage{graphicx}
\usepackage{textcomp}
\usepackage{xcolor}
\usepackage{url} 
\usepackage{multicol}
\usepackage{placeins}
\usepackage{adjustbox} 
\usepackage{makecell} 
\usepackage{arydshln} 
\usepackage{comment}
\usepackage{url}
\usepackage{hyperref}
\def\BibTeX{{\rm B\kern-.05em{\sc i\kern-.025em b}\kern-.08em
    T\kern-.1667em\lower.7ex\hbox{E}\kern-.125emX}}
\begin{document}

\title{Evaluating LLMs and Pre-trained Models for Text Summarization Across Diverse Datasets\\
}

\author{\IEEEauthorblockN{Tohida Rehman}
\IEEEauthorblockA{\textit{Dept. of Information Technology} \\
\textit{Jadavpur University}\\
Kolkata, India\\
tohidarehman.it@jadavpuruniversity.in}
\and
\IEEEauthorblockN{Soumabha Ghosh}
\IEEEauthorblockA{\textit{Dept. of Information Technology} \\
\textit{Jadavpur University}\\
Kolkata, India\\
soumagok@gmail.com}
\and
\IEEEauthorblockN{Kuntal Das}
\IEEEauthorblockA{\textit{Dept. of Information Technology} \\
\textit{Jadavpur University}\\
Kolkata, India\\
iamkuntal2002@gmail.com}
\and
\IEEEauthorblockN{Souvik Bhattacharjee}
\IEEEauthorblockA{\textit{Dept. of Information Technology} \\
\textit{Jadavpur University}\\
Kolkata, India\\
souvikbhattacharjee00076@gmail.com}
\and 
\IEEEauthorblockN{Debarshi Kumar Sanyal}
\IEEEauthorblockA{\textit{School of Mathematical \& Computational Sciences} \\
\textit{Indian Association for the Cultivation of Science}\\
Kolkata, India\\
debarshi.sanyal@iacs.res.in}
\and
\IEEEauthorblockN{Samiran Chattopadhyay}
\IEEEauthorblockA{{\textit{Jadavpur University and}} \\
\textit{Techno India University}\\
Kolkata, India\\
samirancju@gmail.com}
}

\maketitle
\begin{abstract}
Text summarization plays a crucial role in natural language processing by condensing large volumes of text into concise and coherent summaries. As digital content continues to grow rapidly and the demand for effective information retrieval increases, text summarization has become a focal point of research in recent years. This study offers a thorough evaluation of four leading pre-trained and open-source large language models: BART, FLAN-T5, LLaMA-3-8B, and Gemma-7B, across five diverse datasets CNN/DM, Gigaword, News Summary, XSum, and BBC News. The evaluation employs widely recognized automatic metrics, including ROUGE-1, ROUGE-2, ROUGE-L, BERTScore, and METEOR, to assess the models' capabilities in generating coherent and informative summaries. The results reveal the comparative strengths and limitations of these models in processing various text types. 

\end{abstract}


\section{Introduction}
\label{intro}
Text summarization has seen significant advancements, particularly with the rise of deep learning techniques and large pre-trained models. These developments have allowed for more accurate and efficient summarization of lengthy texts, making it a valuable tool for various applications, including content organization, information retrieval, and decision-making. Summarization techniques are generally classified into two categories: extractive and abstractive \cite{el2021automatic}. Extractive methods select key phrases or sentences directly from the source text, while abstractive methods generate concise summaries by rephrasing the original content, similar to how humans would summarize it \cite{el2021automatic}.

\textbf{Contributions.} The primary contributions of this  paper are:
\begin{enumerate}
    \item Fine-tuning and evaluation of four pre-trained models -- BART, FLAN-T5, LLaMA-3-8B, and Gemma-7B -- across five diverse datasets, namely, CNN/DM, Gigaword, News Summary, XSum, and BBC News, for abstractive summarization.
    \item Comparison of summarization quality using established metrics.
    \item Insights into model performance to guide future research and enhancements.
\end{enumerate}
\section{Related Work}
Text summarization in NLP involves creating concise summaries from lengthy documents. Early, methods focused on extractive techniques, selecting key sentences or phrases \cite{el2021automatic}. 

Initially, text summarization research focused on single-document summarization, where key information is extracted from a single document to generate a concise summary. Several recent surveys, such as \cite{koh2022empirical,luo2024comprehensive,zhang2024systematic} provide a comprehensive overview of summarization datasets and techniques, spanning from statistical methods to deep learning models. 

While the early methods struggled with rephrasing and merging content, the introduction of sequence-to-sequence models like the pointer-generator model, augmented with coverage and attention mechanisms \cite{See2017GetTT} marked a significant advancement, although challenges like repeated content and factual inaccuracies or hallucinations persisted. 
Our earlier works \cite{rehman2021automatic,rehman-etal-2022-named,rehman2023research,10172215,rehman2022analysis, rehman2022abstractive,rehman2024analysis} explored the performance of various pre-trained models in open-domain and scholarly domains. Pre-trained models such as T5 \cite{JMLR:v21:20-074}, BART \cite{lewis-etal-2020-bart}, PEGASUS \cite{zhang2020pegasus} and large language models (LLMs) like the GPT-family of models \cite{brown2020language} established new state-of-the-art scores in summarization. This paper chooses models with parameters between a few hundred million to less than 10 billion and assess their performance on a diversity of datasets.

\section{Models Used}
In our approach to summarization, we utilized pre-trained models, fine-tuning them to improve their ability to generate concise and accurate summaries. By adapting both large language models (LLMs) and other pre-trained models, we could compare the performance of diverse models across various datasets. The generated summaries were not only precise but also coherent and informative.

\subsection{BART and FLAN-T5}
For this study, we employed the BART-base model \cite{lewis-etal-2020-bart}, a transformer-based architecture designed for sequence-to-sequence tasks like summarization. BART-base model, with 139 million trainable parameters, uses a denoising autoencoder to reconstruct text, which aids in producing high-quality summaries. Pre-trained on large corpora, the model excels at understanding and generating natural language. 

For this study, we employed the FLAN-T5-base \cite{chung2024scaling} model for the summarization task. The FLAN-T5 tokenizer was then downloaded and applied to prepare the training data. 


\subsection{LLaMA-3-8B and Gemma-7B} 
We fine-tuned the LLaMA-3-8B model \cite{llama3modelcard}, a decoder-only Transformer with 8 billion parameters, trained on publicly available datasets, making it ideal for a wide range of natural language processing tasks. By leveraging LoRA \cite{hu2022lora}, we optimized its performance for each dataset, minimizing both computational costs and memory usage. Similarly, the Gemma-7B model, with 7 billion trainable parameters, was fine-tuned using LoRA optimization techniques for each dataset separately, also minimizing computational costs and memory usage.

\section{Experimental setup}
\subsection{Datasets}
We fine-tuned our models on five distinct datasets, each dataset divided into 1,000 training examples and 100 test examples for concise and precise summaries. We now present a brief overview of the datasets. The {\bf CNN/DM dataset} \cite{hermann2015teaching}, contains over 300k news articles from CNN and the Daily Mail. 
The {\bf Gigaword dataset} \cite{nallapati2016abstractive}, with over 6 million articles from sources like the Associated Press and the New York Times, pairs each article with a headline or brief summary, making it suitable for supervised learning tasks. The {\bf News Summary dataset} includes 4,515 examples with author names, headlines, URLs, short texts, and complete articles, collected from Inshorts, The Hindu, Indian Times, and The Guardian from February to August 2017. The {\bf XSum dataset} \cite{narayan2018don} features 226,711 BBC news articles from 2010 to 2017, each with a one-sentence summary across various domains, such as news, politics, sports, weather, business, technology, science, health, family, education, entertainment, and arts. The {\bf BBC News dataset} \cite{10.1145/1143844.1143892}, on the other hand, comprises 417 political news articles from 2004 to 2005, with each article accompanied by five summaries.

\subsection{Data Processing}
We first obtained the datasets from Hugging Face. 
To prepare the data for fine-tuning BART and FLAN-T5 models, the input articles and their corresponding summaries were organized into separate arrays, which were then paired into tuples to establish a clear relationship between the input text and its target summary. These pairs were subsequently transformed into tensor slices for efficient batch processing during training. This meticulous preprocessing ensures that the BART and FLAN-T5 models learn to generate concise and accurate summaries.

For the fine-tuning of the LLaMA-3-8B and Gemma-7B models, 
we developed a custom function, \texttt{formatting\_prompt\_func()}, to structure the dataset by combining essential components of the main article and its summary into a cohesive string. Each string was marked with an \texttt{EOS\_TOKEN} to indicate the end of the sequence. This structured approach ensures seamless data integration for optimizing LLaMA-3-8B and Gemma-7B for effective text summarization.

\subsection{Implementation Details}
We selected the following pre-trained models from the Hugging Face repository: {BART-base}\footnote{\url{https://huggingface.co/facebook/bart-base}}, {FLAN-T5-base}\footnote{\url{https://huggingface.co/google/flan-t5-base}}, {LLaMA-3-8B}\footnote{\url{https://huggingface.co/unsloth/llama-3-8b-bnb-4bit}}, and {Gemma-7B}\footnote{\url{https://huggingface.co/unsloth/codegemma-7b-bnb-4bit}}.

To fine-tune LLaMA-3-8B and Gemma-7B models, we used a batch size of 2, an evaluation batch size of 1, learning rate of 2e-4, and adaptation matrices with a rank of {\tt r=16}. Additionally, we set {\tt lora\_alpha = 16} as the scaling factor and employed {\tt peft\_config} for loading LoRA. We used the following prompt to generate a summary with LLaMA-3-8B and Gemma-7B LLMs:\\ \texttt{Create a concise and  abstract summary of the following <text>}\\ The same prompt was utilized for both fine-tuning and evaluation. We fine-tuned each model for three epochs on each of the datasets.

For fine-tuning BART and FLAN-T5, we used a batch size of 8, an evaluation batch size of 8, and a learning rate of 5e-5. Across all models, the sequence length during fine-tuning was set to 1000, while the generated outputs had a maximum length of 100. We fine-tuned BART-base and FLAN-T5-base models for three epochs on each of the datasets. 

\subsection{Observation on Computational Resources}
Table \ref{tab:5col7row} presents a summary of the computational resources utilized during the fine-tuning process of the models.

\begin{table}[h]
\centering
\resizebox{\columnwidth}{!}{
\begin{tabular}{|c|c|c|c|c|}
\hline
\textbf{Factors} & \textbf{BART} & \textbf{FLAN-T5} & \textbf{LLaMA-3-8B} & \textbf{Gemma-7B} \\ \hline
Trainable Params & 139M & 248M & 8B & 7B \\ \hline
RAM (12.7GB) & 4.8 & 3.5 & 4 & 4.5 \\ \hline
GPU (15GB) & 8 & 9 & 9.8 & 9.2 \\ \hline
Disk (78GB) & 7.7 & 10.4 & 33.6 & 33.5 \\ \hline
GPU Utilization & 100\% & 97\% & 95\% & 100\% \\ \hline
CPU Utilization & 100\% & 100\% & 96\% & 98\% \\ \hline
Time per Epoch (min) & 6.45 & 1.5 & 50 & 67 \\ \hline
\end{tabular}
}
\caption{Computational Efficiency and Resource Utilization Across Summarization Models.}
\label{tab:5col7row}
\end{table}

\subsection{Evaluation Metrics}
ROUGE (Recall-Oriented Understudy for Gisting Evaluation) \cite{lin2004rouge} is a widely used evaluation metric in NLP tasks, such as summarization, that assesses the quality of a machine-generated summary \( M \) by comparing it to a human-written reference summary \( H \). \textbf{ROUGE-1} measures unigram overlap, \textbf{ROUGE-2} assesses bigram overlap, and \textbf{ROUGE-L} calculates the longest common subsequence (LCS) between \( M \) and \( H \). 
METEOR \cite{banerjee2005meteor} assesses unigram precision and recall, aligning machine-generated and reference summaries at the sentence level with a fragmentation penalty.
BERTScore \cite{zhang2019bertscore} calculates a similarity score for each token in the machine-generated summary \( M \) with each token in the  human-written reference summary \( H \), where token similarity is computed using BERT contextual embeddings. Thus, contextual similarity is used instead of exact token matches. BERTScore is reported to correlate better with human judgments, compared to other metrics \cite{zhang2019bertscore}. 

\section{Results}
\subsection{Model Comparison}
This section compares the performance of the models across the datasets used in this study. We evaluate their performance using multiple metrics: F1 scores for ROUGE-1, ROUGE-2, and ROUGE-L, as well as METEOR and BERTScore. This multi-metric approach provides a comprehensive evaluation of each model's ability to generate high-quality summaries, highlighting their strengths and weaknesses in different scenarios.
\begin{table}[!htbp]
    \centering
    \resizebox{\columnwidth}{!}{
    \begin{tabular}{|p{2.5cm}ccccc|}  
    \hline
    Model Name & R-1 & R-2 & R-L & METEOR & BERTScore \\\hline
    BART & 0.25 & 0.07 & 0.18 & 0.53 & 0.23 \\ \hline
    FLAN-T5 & \textbf{0.34} & \textbf{0.14} & \textbf{0.24} & \textbf{0.58} & \textbf{0.34} \\ \hline
    LLaMA-3-8B & 0.32 & 0.12 & 0.23 & 0.56 & 0.25 \\ \hline
    Gemma-7B &  0.33 & \textbf{0.14} & \textbf{0.24} & \textbf{0.58} & 0.29 \\ \hline
    \end{tabular}
    }
    \caption{Performance Comparison of Models: ROUGE F1-scores, METEOR, and BERTScore on the \textbf{CNN/DM} Dataset.}
    \label{1}
\end{table}

\begin{table}[!htbp]
    \centering
    \resizebox{\columnwidth}{!}{ 
    \begin{tabular}{|p{2.5cm}ccccc|}  
    \hline
    Model Name & R-1 & R-2 & R-L & METEOR & BERTScore \\\hline
    BART & 0.29 & 0.10 & 0.26 & 0.53 & 0.18 \\ \hline
    FLAN-T5 & 0.33 & 0.12 & 0.28 & \textbf{0.57} & \textbf{0.35} \\ \hline
    LLaMA-3-8B & 0.29 & 0.09 & 0.25 & 0.50 & 0.25 \\ \hline
    Gemma-7B & \textbf{0.35} & \textbf{0.14} & \textbf{0.32} & 0.55 & 0.29 \\ \hline
    \end{tabular}
    }
    \caption{Performance Comparison of Models: ROUGE F1-scores, METEOR, and BERTScore on the \textbf{Gigaword} Dataset.}
    \label{2}
\end{table}

\begin{table}[!htbp]
    \centering
    \resizebox{\columnwidth}{!}{ 
    \begin{tabular}{|p{3cm}ccccc|}  \hline
    Model Name & R-1 & R-2 & R-L & METEOR & BERTScore \\\hline
    BART & \textbf{0.56} & \textbf{0.36} & \textbf{0.45} & \textbf{0.73} & \textbf{0.49} \\ \hline
    FLAN-T5 & 0.52 & 0.29 & 0.39 & 0.70 & 0.44 \\ \hline
    LLaMA-3-8B & 0.51 & 0.27 & 0.37 & 0.67 & 0.41 \\ \hline
    Gemma-7B & 0.51 & 0.28 & 0.38 & 0.68 & 0.38 \\ \hline
    \end{tabular}
    }
    \caption{Performance Comparison of Models: ROUGE F1-scores, METEOR, and BERTScore on the  \textbf{News Summary} dataset.}
    \label{3}
\end{table}

\begin{table}[!htbp]
    \centering
    \resizebox{\columnwidth}{!}{ 
    \begin{tabular}{|p{3cm}ccccc|}  \hline
    Model Name & R-1 & R-2 & R-L & METEOR & BERTScore \\\hline
    BART & 0.27 & 0.07 & 0.21 & 0.57 & 0.22 \\ \hline
    FLAN-T5 & 0.35 & 0.13 & 0.27 & \textbf{0.61} & \textbf{0.30} \\ \hline
    LLaMA-3-8B & 0.37 & 0.15 & 0.29 & 0.56 & 0.27 \\ \hline
    Gemma-7B & \textbf{0.39} & \textbf{0.18} & \textbf{0.32} & \textbf{0.61} & \textbf{0.30} \\ \hline
    \end{tabular}
    }
    \caption{Performance Comparison of Models: ROUGE F1-scores, METEOR, and BERTScore on the \textbf{XSum} dataset.}
    \label{4}
\end{table}

\begin{table}[!htbp]
    \centering
    \resizebox{\columnwidth}{!}{ 
    \begin{tabular}{|p{3cm}ccccc|}  \hline
    Model Name & R-1 & R-2 & R-L & METEOR & BERTScore \\\hline
    BART & \textbf{0.61} & 0.47 & \textbf{0.42} & \textbf{0.75} & \textbf{0.51} \\ \hline
    FLAN-T5 & 0.27 & 0.23 & 0.25 & 0.64 & 0.15 \\ \hline
    LLaMA-3-8B & \textbf{0.61} & \textbf{0.49} & 0.40 & 0.74 & \textbf{0.51} \\ \hline
    Gemma-7B & 0.51 & 0.39 & 0.36 & 0.71 & 0.33 \\ \hline
    \end{tabular}
    }
    \caption{Performance Comparison of Models: ROUGE F1-scores, METEOR, and BERTScore on the \textbf{BBC News} dataset.}
    \label{5}
\end{table}
\begin{enumerate}
\item \textbf{Dataset: CNN/DM}\\
The results for ROUGE-1 (R-1), ROUGE-2 (R-2), ROUGE-L (R-L), METEOR, and BERTScore on the CNN/DM dataset are presented in Table \ref{1}. It can be observed that, among the four models, the fine-tuned FLAN-T5 model outperforms the others, achieving the highest values for the ROUGE, BERTScore, and METEOR metrics. Furthermore, the fine-tuned Gemma-7B model demonstrates performance similar to that of the fine-tuned FLAN-T5 model, particularly in terms of ROUGE-2, ROUGE-L, and METEOR scores.

\item \textbf{Dataset: Gigaword} \\
The results for ROUGE-1 (R-1), ROUGE-2 (R-2), ROUGE-L (R-L), BERTScore, and METEOR on the Gigaword dataset are presented in Table \ref{2}. It is evident from the results that the fine-tuned Gemma-7B model achieves the highest scores for the ROUGE metrics. In contrast, the fine-tuned FLAN-T5 model outperforms the others in terms of BERTScore and METEOR values.

\item \textbf{Dataset: News Summary}\\
The results for ROUGE-1 (R-1), ROUGE-2 (R-2), ROUGE-L (R-L), BERTScore, and METEOR on the News Summary dataset are presented in Table \ref{3}. Among the four models, the fine-tuned BART model achieves the highest scores across all evaluated metrics, namely, ROUGE, BERTScore, and METEOR.

\item \textbf{Dataset: XSum} \\
The results for ROUGE-1 (R-1), ROUGE-2 (R-2), ROUGE-L (R-L), BERTScore, and METEOR on the XSum dataset are presented in Table \ref{4}. The fine-tuned Gemma-7B model demonstrates superior performance, achieving the highest scores across all evaluated metrics, namely, ROUGE, BERTScore, and METEOR. Furthermore, the fine-tuned FLAN-T5 model demonstrates performance similar to that of the fine-tuned Gemma-7B model, particularly in terms of METEOR score and BERTScore.

\item \textbf{Dataset: BBC News} \\
The results for ROUGE-1 (R-1), ROUGE-2 (R-2), ROUGE-L (R-L), BERTScore, and METEOR on the BBC News dataset are displayed in Table \ref{5}. Among the four models, the fine-tuned BART model achieves the highest scores for ROUGE-1, ROUGE-L, BERTScore, and METEOR. However, the fine-tuned LLaMA-3-8B model outperforms others in ROUGE-2 and attains the same highest scores as obtained by BART for R-1 and BERTScore. It is also very close to BART in R-L and METEOR.
\end{enumerate}
We also performed an evaluation of the summaries using ChatGPT\footnote{\url{https://chatgpt.com/}} as a judge. In particular, we randomly selected 5 examples from each dataset and collected the golden as well as the four model-generated summaries for each example. Then, we prompted ChatGPT to select the most preferred machine generated summary given the original text,  the golden summary and the four generated ones. Ranking the models by the number of times their summaries were most preferred, we found that Gemma-7B was the top choice for Gigaword and XSum, LLaMA-3-8B for BBC News, and BART for News Summary. For CNN/DM, LLaMA-3-8B was preferred in 2 out of 5 cases; in the remaining 3 cases, a different model was preferred in each instance. Except for CNN/DM, the above preferences align closely with the earlier automatic metric-based evaluation. A major issue with BART and FLAN-T5 was that their outputs sometimes contained repeated information and grammatical errors, which were rarely present in LLaMA-3-8B and Gemma-7B. Another concern is that hallucinations \cite{rehman2023hallucination} occasionally appeared in the generated summaries.

\subsection{Case study}
We will now showcase various examples illustrating the results generated by the models discussed in this paper. In each case study presented, errors or incorrect information are highlighted in \colorbox{magenta!30}{magenta}, reduandant information is shown in \colorbox{cyan!10}{cyan}, and  repeated words are indicated in \colorbox{yellow!50}{yellow}.

\begin{figure*}[!htpb] 
\centering 
\small
\renewcommand{\arraystretch}{1.7} 
\begin{tabular}{ |p{17.2cm}|} \hline
        \textbf{Ground-truth summary:} 
        ``Fire breaks out at the General Electric Appliance Park in Louisville, Kentucky. City officials report no one is believed to be injured or trapped." \\
        \hline
        \textbf{BART:} ``Fire damages industrial park in Kentucky, official says. Video shows blaze spreading thick smoke over area. No injuries or trapped, official says. \colorbox{cyan!10}{Officials didn't know what had caused the fire, which had gone to at least four alarms.} \colorbox{yellow!50}{Officials didn't know what had caused the fire.}"\\
        \hline
        \textbf{FLAN-T5:} ``Fire breaks out in Kentucky industrial park, sending plumes of thick smoke over the area. No reports of anyone injured or trapped in the blaze, officials say. \colorbox{cyan!10}{GE's facility in Louisville Appliance Park is ``revitalizing manufacturing in the United States".}" \\
        \hline
        \textbf{LLaMA-3-8B:} ``\colorbox{magenta!30}{NEW:} Fire began shortly before 7 a.m. at General Electric Appliance Park. \colorbox{magenta!30}{NEW:} No reports of anyone injured or trapped, official says. \colorbox{magenta!30}{NEW:} \colorbox{cyan!10}{Firefighters spray water from the periphery of affected buildings.} \colorbox{magenta!30}{NEW:} \colorbox{cyan!10}{Park is large, such that 34 football fields could fit in one of its warehouses.}" \\
        \hline
        \textbf{Gemma-7B:} ``Massive fire breaks out at General Electric Appliance Park in Louisville, Kentucky. \colorbox{cyan!10}{Firefighters take up positions around affected buildings, spraying water. Authorities don't know what caused the fire, which} \colorbox{cyan!10}{has gone to at least four alarms.}" \\
        \hline
    \end{tabular}
    \caption{Ground-truth summary and summaries generated by different models on the CNN/DM test dataset. Input taken from \url{https://edition.cnn.com/2015/04/03/us/kentucky-ge-fire/index.html}.}
    \label{fig:Casestudy-cnndm} 
\end{figure*}

\begin{figure*}[!htpb] 
\centering 
\small
\renewcommand{\arraystretch}{1.7} 
\begin{tabular}{ |p{17.2cm}|} \hline
\textbf{Ground-truth summary}  ``Sri lanka closes schools as war escalates."\\
\hline  
\textbf{BART generated:} ``Sri lanka government shuts government schools amid military campaign against rebels."\\
\hline 
\textbf{FLAN-T5 generated:} ``Sri lanka announces closure of government schools as tamil separatists campaign intensifies in north of \colorbox{magenta!30}{london}."\\
\hline 
\textbf{LLaMA-3-8B generated:} ``Srilanka shuts schools as war intensifies." \\
\hline 
\textbf{Gemma-7B generated:} ``Sri lanka closes schools amid escalating war."\\
\hline
\end{tabular}
\caption{Ground-truth summary and summaries generated by different models on the Gigaword dataset. Input taken from \url{https://www.cfr.org/backgrounder/sri-lankan-conflict}}
\label{fig:Casestudy-gigaword} 
\end{figure*}

In the case study shown in Figure \ref{fig:Casestudy-cnndm}, the fine-tuned BART model makes a mistake by repeating the phrase ``Officials didn’t know what had caused the fire,'' which had already been stated earlier in the summary, leading to redundancy. The fine-tuned LLaMA-3-8B model makes mistakes by unnecessarily repeating ``NEW:" before relevant details and including extraneous information, such as the size of the park, which is irrelevant to the main summary.
In the case study in Figure \ref{fig:Casestudy-gigaword}, the fine-tuned FLAN-T5 model introduces an error by mistakenly inserting ``London" into the summary, which is contextually irrelevant, whereas the other models produce more accurate and relevant summaries. 



\section{Conclusion}
This paper provides a comprehensive analysis of various pre-trained models and large language models for abstractive text generation, evaluating their ability to produce coherent, informative, and concise summaries. While these models show promising results, there are still areas for improvement, especially in syntactical accuracy and factual consistency. Future work will include human evaluations or a more detailed evaluation with LLM-as-a-judge to assess the quality of AI-generated summaries. Additionally, we plan to analyze the issues of hallucination and its mitigation strategies in the context of abstractive summarization.

\bibliographystyle{unsrt}
\bibliography{llm_ref,anthology}
\end{document}